%% file: ms.tex
\documentclass[10pt,twocolumn,letterpaper]{article}

\usepackage{cvpr}
\usepackage{times}
\usepackage{epsfig}
\usepackage{graphicx}
\usepackage{amsmath}
\usepackage{amssymb}
\usepackage{xcolor}
\usepackage{multirow}
\usepackage{makecell}
\usepackage{url}
\usepackage{caption}
\usepackage{algorithm}
\usepackage{algorithmicx}
\usepackage{algpseudocode}
\usepackage{amsmath}
\usepackage{booktabs}
\usepackage{appendix}

\usepackage[pagebackref=true,breaklinks=true,letterpaper=true,colorlinks,bookmarks=false]{hyperref}

\cvprfinalcopy 

\author{
  Fanbo Xiang$^1$ \quad Yuzhe Qin$^1$ \quad Kaichun Mo$^2$ \quad Yikuan
  Xia$^1$ \quad Hao Zhu$^1$ \\ Fangchen Liu$^1$ \quad
  Minghua Liu$^1$ \quad Hanxiao Jiang$^3$ \quad Yifu Yuan$^5$ \quad He Wang$^2$ \quad Li Yi$^4$ \\
  \quad Angel X. Chang$^3$ \quad Leonidas Guibas$^2$ \quad Hao Su$^1$\\
  $^1$UC San Diego 
  $^2$Stanford University 
  $^3$Simon Fraser University 
  $^4$Google Research 
  $^5$UC Los Angeles\\
  \href{https://sapien.ucsd.edu}{https://sapien.ucsd.edu} 
}

\ifcvprfinal\fi
\begin{document}

\title{SAPIEN: A SimulAted Part-based Interactive ENvironment}

\maketitle 

\label{abs}
\input{tex/abstract.tex}

\section{Introduction}
\label{sec:intro}
\input{tex/intro_new.tex}

\section{Related Work}
\label{sec:related}
\input{tex/related.tex}

\section{SAPIEN Simulation Environment}
\label{sec:simulator}
\input{tex/simulator.tex}


\section{Tasks and Benchmarks}
\label{sec:exp}
\input{tex/exp.tex}

\section{Conclusion}
\label{sec:conclusion}
\input{tex/conclusion.tex}

\section*{Acknowledgements}
\label{sec:ack}
\input{tex/ack.tex}

{\small
\bibliographystyle{ieee_fullname}
\bibliography{ms}
}

\end{document}


\title{SAPIEN: a SimulAted Part-based Interactive ENvironment\\Supplementary Material}

\author{
  Fanbo Xiang$^1$ \quad Yuzhe Qin$^1$ \quad Kaichun Mo$^2$ \quad Yikuan
  Xia$^1$ \quad Hao Zhu$^1$ \\ Fangchen Liu$^1$ \quad
  Minghua Liu$^1$ \quad Hanxiao Jiang$^3$ \quad Yifu Yuan$^5$ \quad He Wang$^2$ \quad Li Yi$^4$ \\
  \quad Angel X. Chang$^3$ \quad Leonidas Guibas$^2$ \quad Hao Su$^1$\\
  $^1$UC San Diego 
  $^2$Stanford University 
  $^3$Simon Fraser University 
  $^4$Google Research 
  $^5$UC Los Angeles\\
  \href{https://sapien.ucsd.edu}{https://sapien.ucsd.edu} 
}

\makeatletter
\let\@oldmaketitle\@maketitle
\renewcommand{\@maketitle}{\@oldmaketitle
\centering
  \includegraphics[width=0.9\linewidth]{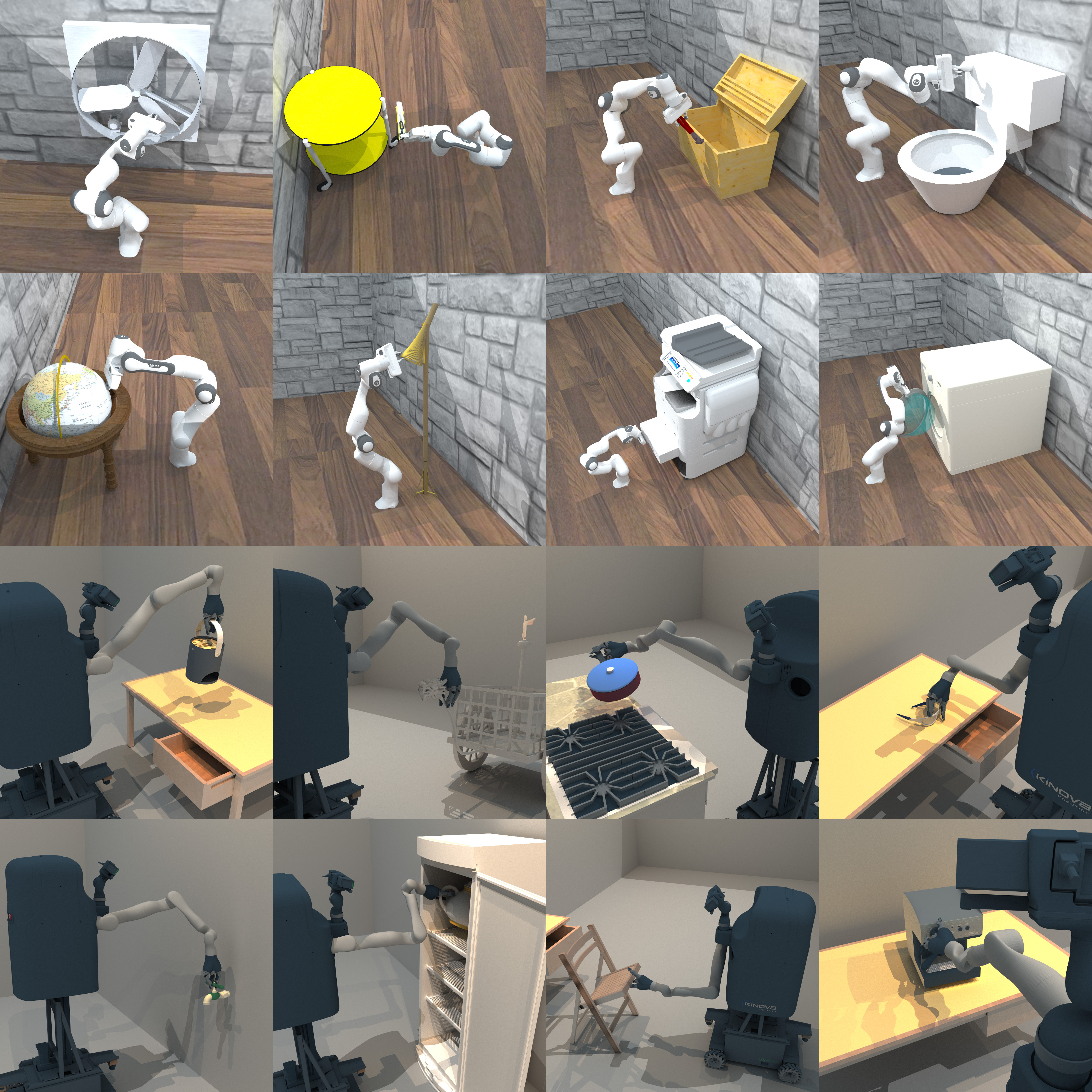}
    \captionof{figure}{\textbf{Diverse manipulation tasks supported by SAPIEN}\label{fig:teaser}}
    \bigskip}
\makeatother

\maketitle

\appendix

\section*{Table of Contents}
\begin{itemize}
\item \textbf{Appendix A} Details on PartNet-Mobility Annotation System.
\item \textbf{Appendix B} Experiment details on movable part segmentation and motion recognition tasks.
\item \textbf{Appendix C} Terminologies
\end{itemize}

\section*{Appendix A: Annotation System}
\label{appB:annotation}
\input{supp_tex/appBanno.tex}
\section*{Appendix B: Movable Part Segmentation and Motion Recognition}
\label{appC:net}
\input{supp_tex/appCnet.tex}
\section*{Appendix C: Terminology}
\label{appD:terminology}
\input{supp_tex/appDwrapper.tex}


{\small
\bibliographystyle{ieee_fullname}
\bibliography{supplement}
}

%% file: tex/abstract.tex
\begin{abstract}
Building home assistant robots has long been a pursuit for vision and robotics researchers. To achieve this task, a simulated environment with physically realistic simulation, sufficient articulated objects, and transferability to the real robot is indispensable. Existing environments achieve these requirements for robotics simulation with different levels of simplification and focus. We take one step further in constructing an environment that supports household tasks for training robot learning algorithm. Our work, SAPIEN, is a realistic and physics-rich simulated environment that hosts a large-scale set for articulated objects. Our SAPIEN enables various robotic vision and interaction tasks that require detailed part-level understanding.We evaluate state-of-the-art vision algorithms for part detection and motion attribute recognition as well as demonstrate robotic interaction tasks using heuristic approaches and reinforcement learning algorithms. We hope that our SAPIEN can open a lot of research directions yet to be explored, including learning cognition through interaction, part motion discovery, and construction of robotics-ready simulated game environment.
\end{abstract}

%% file: tex/intro_new.tex

\begin{figure}[t]
    \centering
    \includegraphics[width=\linewidth]{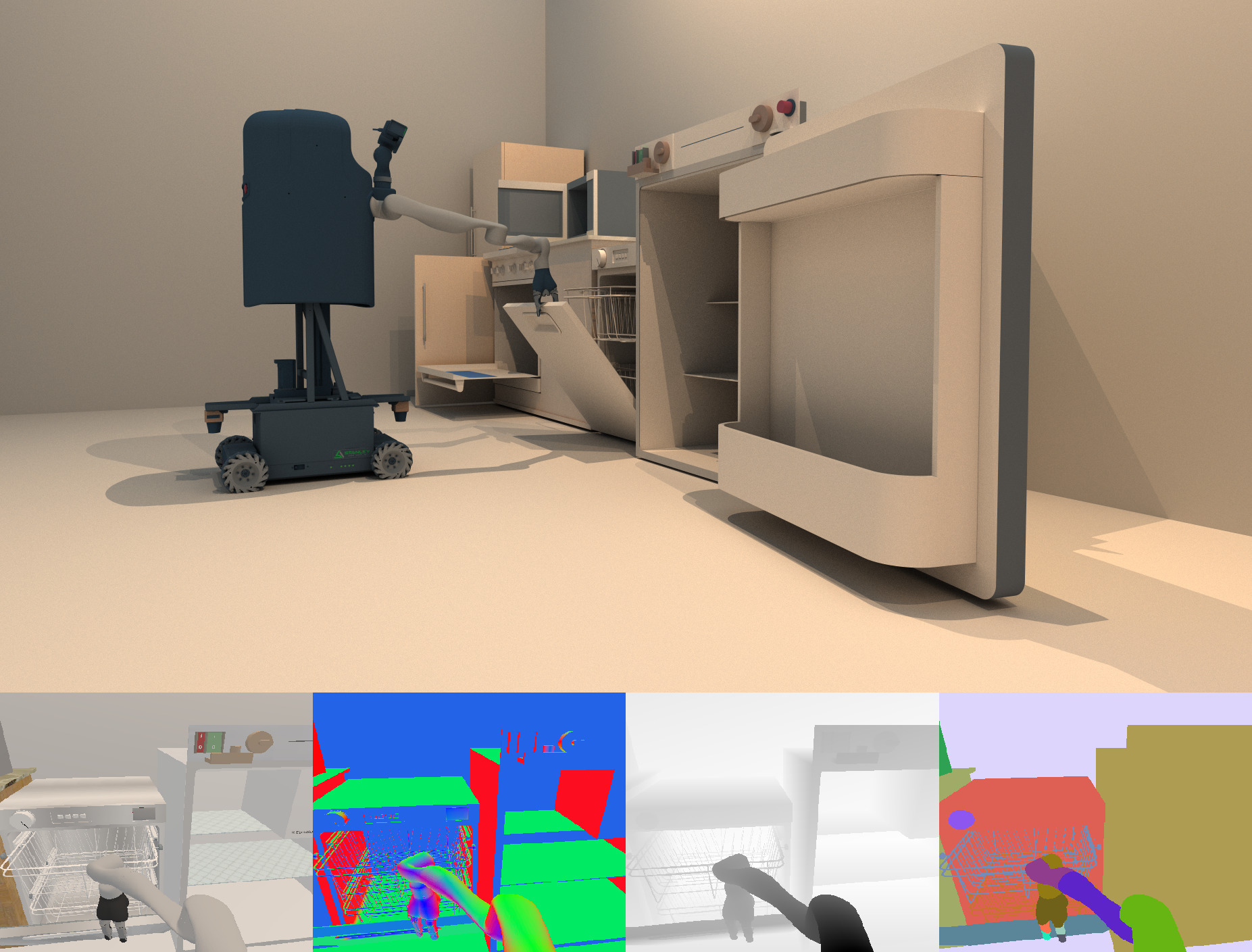}
    \caption{\textbf{Robot-object Interaction in SAPIEN.} We show the ray-traced scene (top) and robot camera views (bottom): RGB image, surface normals, depth and semantic segmentation of motion parts, while a robot is learning to operate a dishwasher.}
    \label{fig:teaser}
    \vspace{-4mm}
\end{figure}

\begin{table*}[t]
\centering
\setlength{\tabcolsep}{3pt}
\begin{tabular}{l|l|l|l|l|l}
\toprule
Environment   & Level  & Physics  & Rendering  & Tasks  & Interface \\ \midrule
Habitat~\cite{habitat19iccv}* & Scene & Static+ & Real Photo & Navigation, Vision & Python, C++ \\
AI2-THOR~\cite{ai2thor}* & Scene-Object & Dynamic & Unity & Navigation+, Vision & Python, Unity \\
\small{OpenAI Gym MuJoCo}~\cite{brockman2016openaigym} & Scene-Object & Dynamic & OpenGL(fixed)  & Learning, Robotics          & Python    \\
RLBench\cite{james2019rlbench} &  Scene-Object & Dynamic & V-REP, PyRep    & \small{Learning, Vision, Robotics} & Python, V-REP    \\ \midrule
SAPIEN &  \small{Scene-Object-Part} & Dynamic & Customizable & \small{Learning, Vision, Robotics} & Python, C++         \\
\bottomrule
\end{tabular}
\caption{\textbf{Comparison to other Simulation Environments.} Habitat~\cite{habitat19iccv} is a representative for navigation environments, which include Gibson~\cite{xia2018gibson,xia2019interactive}, Minos~\cite{savva2017minos}; they primarily use static physics but are starting to add interactions very recently. AI2-THOR~\cite{ai2thor} is a representative for game-like interactive environments; these environments usually support navigation with limited object interactions. OpenAI Gym~\cite{brockman2016openaigym} and RLBench~\cite{james2019rlbench} provide interactive environments, but the use of commercial software limits their customizability.}
\vspace{-4mm}
\label{tab: environments}
\end{table*}

To achieve human-level perception and interaction with the 3D world, home-assistant robots must have the capability to use perception to interact with 3D objects~\cite{das2018embodied,yang2019embodied}. For a robot to help put away groceries, it must be able to open the refrigerator by locating the door handle, pulling the door and fetching the target objects. 

One direct way to address the problem is to train robots by interacting with the real environment~\cite{mandlekar2018roboturk,calli2017yale, levine2018learning}. However, training robots in the real world could be very time consuming, costly, unstable, and potentially unsafe. Moreover, a slight perturbation in hardware or environment setup can result in different outcomes in the real world, thus inhibiting reproducible research. Researchers, therefore, have long been pursuing simulated environments for tasks such as navigation~ \cite{savva2017minos,xia2018gibson,habitat19iccv,mattersim,brodeur2017home,wu2018building,gupta2017cognitive,xia2018gibson} and control~\cite{gazebo,vrep,todorov2012mujoco,coumans2016pybullet}.

Constructing simulated environments for robot learning with transferability to the real world is a non-trivial task. It faces challenges from four major aspects: 1) The environment needs to reproduce the real-world physics to some level. As it is still infeasible to simulate real-world physics exactly, any physical simulator needs to decide the level-of-details and accuracy it operates on. Some approximate physics by simulating rigid bodies and joints\cite{physx, todorov2012mujoco, coumans2016pybullet}; some handle soft deformable objects~\cite{todorov2012mujoco, coumans2016pybullet}; and others simulate fluid~\cite{todorov2012mujoco,schenck2018spnets}. 2) The environment should incorporate the simulation of real robots, being able to reproduce the behaviors of real robotics manipulators, sensors and controllers~\cite{murali2019pyrobot}. Only this can enable seamless transfer to the real-world after training. 3) The environment needs to produce physically accurate renderings to mitigate the visual domain gap. 
4) Most importantly, the environment requires sufficient content, scenes and objects for the robot to interact with, since data diversity is always critical for training and evaluating learning-based algorithms. The content also determines how much we shall address challenges in the previous tasks: data with soft objects such as cloth requires deformable body simulation; translucent objects require special rendering techniques, and specific robot requires a specific interface.

Existing environments achieve these requirements for robotics simulation with different levels of simplification and focus. For example, OpenAI Gym~\cite{brockman2016openaigym} provides an interactive and easy-to-use interface; Gibson~\cite{xia2018gibson} and AI Habitat~\cite{habitat19iccv} use photorealistic rendering for semantic navigation tasks. A more detailed discussion of popular environment features can be found in Sec~\ref{sec:related}. These environments can support the benchmarking and training of down-stream tasks such as navigation, low-level control, and grasping. However, from the perspective of tasks, there still lacks environments that target at object manipulation of daily objects, a basic skill of household robots. In a household environment, a great portion of daily objects are articulated and require manipulation: bottles with caps, ovens with doors, electronics with switches and buttons. Notably, RLBench~\cite{james2019rlbench} (unpublished) provides well-defined robotics tasks and realistic controller interface with detailed manipulation demonstration, but it lacks diversity in its simulated scenarios. 

We take one step further in constructing an environment that supports the manipulation of diverse articulated objects. 
Our system, SAPIEN, is a realistic and physics-rich simulated environment that hosts a large set for articulated objects. At the core of SAPIEN are three main components: 1) SAPIEN Engine, an interaction-rich and physics-realistic simulation environment integrating PhysX physical engine and ROS control interface; this engine supports accurate simulation of rigid body and joint constraints for simulation of articulated objects. 2) SAPIEN Asset, including PartNet-Mobility dataset, which contains 14K movable parts over 2,346 3D articulated models from 46 common indoor object categories, richly annotated with kinematic part motions and dynamic interactive attributes; 3) SAPIEN Renderer, with both fast-frame-rate OpenGL rasterizer and more photorealistic ray-tracing options. We demonstrate that our SAPIEN enables a large variety of robotic perception and interaction tasks by benchmarking state-of-the-art vision algorithms for part detection and motion attribute recognition. We also show a variety of robotic interaction tasks that SAPIEN supports by demonstrating heuristic approaches and reinforcement learning algorithms.

%% file: tex/related.tex
\renewcommand{\thefootnote}{*}

\vspace{-1mm}
\paragraph{Simulation Environments.} 
In recent years, there has been a proliferation of indoor simulation environments primarily designed for navigation, visual recognition and reasoning~\cite{savva2017minos,wu2018building,xia2018gibson,habitat19iccv,mattersim}. Static environments, based on synthetic scenes~\cite{wu2018building} or real-world RGB-D scans~\cite{mattersim} and  reconstructions~\cite{savva2017minos,xia2018gibson,habitat19iccv}, are able to provide images that closely resembles reality, minimizing the domain gap in the visual aspect. However, they usually offer very limited or no object interactions, failing to capture the dynamic and interactive nature of the real world. 



In order to allow for more interactive features to the environment, researchers leverage partial functionalities of game engines or physics engine to provide photorealistic rendering together with interactions~\cite{qiu2017unrealcv, muller2018sim4cv,ai2thor,puig2018virtualhome,brodeur2017home,yan2018chalet,VRKitchen}. When agents interact with objects in these environments, it is via high level state changes triggered by explicit commands (\eg ``open refrigerator''), or proximity (\eg refrigerator door opens when the robot or robot arm is next to the trigger region). In addition, the underlying physics is often over-simplified such as direct exertion of force and torque. While they enable research on high-level object interactions, they cannot close the gap between high-level instructions and the low-level dynamics for not including accurate simulation of articulated robots and objects by design. This limits the use of such simulators for learning of detailed low-level robot-object interactions.  

Finally, there are environments that integrate full-featured physics engines. These environments are favored in continuous control and reinforcement learning tasks. OpenAI Gym~\cite{brockman2016openaigym}, RLLAB~\cite{duan2016benchmarking}, DeepMind Control Suite~\cite{deepmindcontrolsuite2018} and DoorGym~\cite{urakami2019doorgym} integrate MuJoCo physical engine to provide RL environments. Arena~\cite{arena}, a platform that supports multi-agent environments, is built on top of Unity~\cite{juliani2018unity}. PyBullet~\cite{coumans2016pybullet}, a real-time physics engine with Python interface, powers a series of projects focusing on robotics tasks~\cite{zeng2019tossingbot,lutter2019deep}. Gazebo ~\cite{gazebo}, a high-level visualization and modeling package, is widely used in robotics community ~\cite{meyer2012comprehensive, hugues2006simbad}. Recently, RLBench~\cite{james2019rlbench}, a benchmark and physical environment for robot learning, uses V-REP~\cite{vrep} as the backend to provide diverse tasks for robot manipulation. Our environment, SAPIEN engine, is directly based on the open-source Nvidia PhysX API~\cite{physx}, which has comparable performance and interface with PyBullet, avoiding the unnecessary complication introduced by game engine infrastructures, or any barriers from commercial software such as MuJoCo and V-REP. Table \ref{tab: environments} provides a brief summary of several representative environments.

One bottleneck of these robotic simulators is their limited rendering capability, which causes a gap between simulation and the real world.
Another constraint of many of these environments, including RLBench~\cite{james2019rlbench} and DoorGym~\cite{urakami2019doorgym}, is that they are very task-centric, designed to work for only a few predefined tasks. Our SAPIEN simulator, equipped with 2,346 3D interactive models from 46 object categories and flexible rendering pipelines, provides robot agents a virtual environment for learning a large set of complex, diverse and customizable robotic interaction tasks. 
\begin{table}[t]
\centering
\setlength{\tabcolsep}{3pt}
\begin{tabular}{r|r|r|r}
\toprule
Dataset & \#Categories & \#Models & \#Motion Parts \\ \midrule
Shape2Motion\cite{wang2019shape2motion} & 45 & 2,440 & 6,762 \\
RPM-Net\cite{yan19rpmnet} & 43 & 949 & 1,420 \\
Hu \etal \cite{hu2017learning} & - & 368 & 368 \\
RBO*\cite{martin2019rbo} & 14 & 14 & 21 \\ \midrule
\textbf{Ours} & 46 & 2,346 & \textbf{14,068} \\
\bottomrule
\end{tabular}
\caption{\textbf{Comparison of Articulated Part Datasets.} *RBO is collected in real-world with long video sequences. }
\label{tab:motion_datasets}
\vspace{-3mm}
\end{table}

\vspace{-3mm}
\paragraph{Simulation Content.} 
Navigation environments typically use datasets providing real-world RGB-D scans~\cite{xia2018gibson, Matterport3D, replica19arxiv}, and/or high-quality synthetic scenes~\cite{song2017ssc}. 
Simulation environments that leverage game engines~\cite{qiu2017unrealcv, muller2018sim4cv, brodeur2017home, VRKitchen, ai2thor} come with manually designed or procedurally generated game scenes. For environments with detailed physics and reinforcement learning support~\cite{brockman2016openaigym, deepmindcontrolsuite2018, duan2016benchmarking}, they usually support very few scenarios with simple objects and robot agents. Notably, RLBench~\cite{james2019rlbench} provides a relatively large robot learning dataset with varied tasks. To address the lack-of-content problem, our work provides a large-scale simulation-ready dataset, PartNet-Mobility dataset, that is constructed from 3D model datasets including PartNet~\cite{Mo_2019_CVPR} and ShapeNet~\cite{chang2015shapenet}.





There are also shape part datasets with part articulation annotations.
Table~\ref{tab:motion_datasets} summarizes recent part mobility datasets. 
The RBO dataset~\cite{martin2019rbo} is a collection of 358 RGB-D video sequences of humans manipulating 14 objects which are reconstructed as articulated 3D meshes. The meshes have detailed part motion parameters and have realistic textures. 
Other datasets annotate 3D synthetic CAD models with articulation information.
Hu \etal~\cite{hu2017learning} introduced a dataset of 368 mobility part articulations with diverse types. RPM-Net~\cite{yan19rpmnet} provides another dataset with 969 objects and 1,420 mobility units.
Shape2Motion~\cite{wang2019shape2motion} provides a dataset of 2,440 objects and 6,762 movable parts for mobility analysis, but it does not provide RGB textures and motion limits that hinders physical simulation. 
Compared to these datasets, our dataset contains comparable number of objects (2,346), but with much more movable part annotations (14,068). Besides, our models have textures and motion range limits, which are crucial for the dataset to be simulatable.

%% file: tex/simulator.tex
\begin{figure*}[t]
    \centering
    \includegraphics[width=0.9\linewidth]{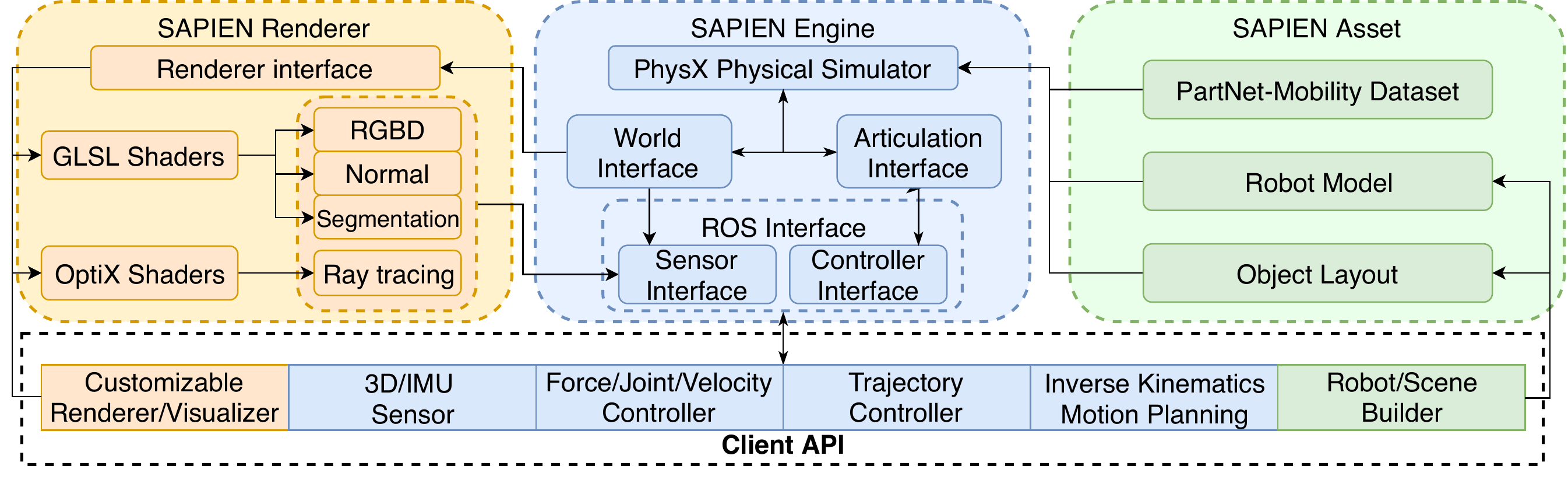}
    \caption{\textbf{SAPIEN Simulator Overview.} The left box shows SAPIEN Renderer, which takes custom shaders and scene information to produce images such as RGB-D and segmentation. The middle box shows SAPIEN Engine, which integrates PhysX simulator and ROS control interface that enables various robot actions and utilities. The right box shows SAPIEN Asset, which contains the large-scale PartNet-Mobility dataset that provides simulatable models with part-level mobility.
    }
    \label{fig:sapien_diagram}
    \vspace{-3mm}
\end{figure*}
SAPIEN aims to integrate state-of-the-art physical simulators, modern graphics rendering engines, and user-friendly robotic interfaces into a unified framework (Figure~\ref{fig:sapien_diagram}), to support a diverse set of robotic perception and interaction tasks.
We develop the environment with C++ for efficiency and provide Python wrapper API for ease-of-use at the user end.
Below we detailedly introduce the three main components: SAPIEN engine, SAPIEN asset and SAPIEN renderer.

\subsection{SAPIEN Engine}
We use the open-source Nvidia PhysX physical engine to provide detailed robot-object interaction simulation.
The system provides Robot Operating System (ROS) supports that are easy-to-use for end-stream robotic research. 
We provide both synchronous and asynchronous modes of simulation to support reinforcement learning training and robotics tasks.


\vspace{-3mm}
\paragraph{Physical Simulation.}
We choose PhysX 4.1~\cite{physx} to provide rigid body kinematics and dynamics simulation, since it is open-source, simplistic, and provides functionalities designed for robotics. To simulate articulated bodies, we provide 3 different body-joint systems: kinematic joint system, dynamic joint system, and PhysX articulation. 
The kinematic joint system provides kinematic objects with parent-child relations, suitable for simulating very heavy objects that are not affected by small forces. 
Dynamic joint systems use PhysX joints to drive rigid bodies towards constraints, suitable for simulating complicated objects that do not require accurate control. 
PhysX articulation is a system specifically designed for robot simulation. 
It natively supports accurate force control, P-D control and inverse dynamics with the cost of relatively low speed.

\vspace{-3mm}
\paragraph{ROS Interfaces.}
 Robot Operating System (ROS)~\cite{quigley2009ros} is a generic and widely-used framework for building robot applications. Our ROS interface bridges the gap between ROS and physical simulator, as well as provides a set of high-level APIs for interacting with robots in the physics world. It supports three levels of abstractions: direct force control, ROS controllers and motion planning interface.
 
 In the lowest level control, forces and torques are directly applied on joints, similar to OpenAI Gym~\cite{brockman2016openaigym}. This control method is simple and intuitive, but rather difficult to transfer to real environments, since real-world dynamics are quite different from the simulated ones, and the continuous nature present in real-robots are fundamentally different from the discretized approximation in simulations. 
 For high-level control, we provide joint space and Cartesian coordinate space control APIs. We build various controllers (Figure~\ref{fig:sapien_diagram}) based upon~\cite{chitta2017ros_control} and implement standard interface. A typical use case is to move the robot arm to a desired 6-DoF pose with specific path constraints. Thus, at the highest level, we provide motion planning support based on the popular MoveIt framework~\cite{chitta2012moveit}, which can generate motion plans that effectively move the robot around without collision. 

\vspace{-3mm}
\paragraph{Synchronous and Asynchronous Modes.}
Our SAPIEN Engine (see Figure~\ref{fig:sapien_diagram} middle) can support both synchronous and asynchronous simulation modes. In synchronous mode, the simulation step is controlled by the client, which is common in training reinforcement learning algorithm~\cite{brockman2016openaigym}. For example, the agent receives observations from simulated environments and uses a customized policy model, often a neural network, to generate the corresponding action. Then the simulation runs forward for a step. In this synchronous mode, the simulation and client algorithms are integrated together.

However, for real-world robotics, the simulation and client response need to be asynchronous~\cite{gazebo} and separated. The simulation should run independently, like the real world, while the client uses the same API as a real robot to interact with the simulation backend. To build such a framework, we create multiple sensors and controllers following the ROS API. After simulation starts, the client receives information from sensors and uses the controller interface (see Figure~\ref{fig:sapien_diagram}) to command robots via TCP/IP communication. The timestamp is synchronized from simulation to the client side, acting as a proxy for the real-world clock time. Under the framework, the simulated robots can use the same code as their real counterparts because most real robot controllers and sensors have exactly the same interface as our simulator API. This provides one important advantage: it enables robot researchers to migrate between simulated robots and real robots without any extra setup. 

\begin{figure*}[!h]
    \centering
    \includegraphics[width=\linewidth]{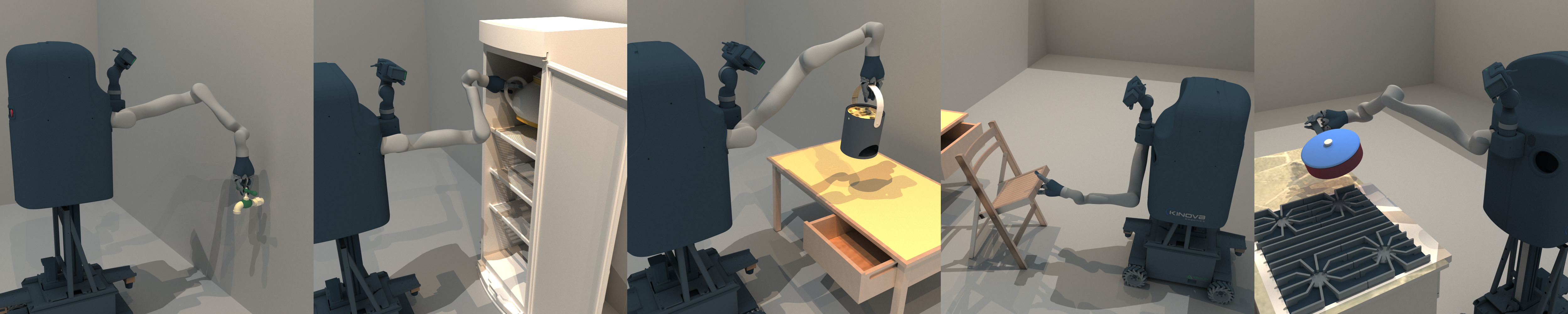}
    \caption{\textbf{SAPIEN Enables Many Robotic Interaction Tasks.} From left to right, we show five examples: faucet manipulation, object fetching, object lifting, chair folding, and object placing.}
    \label{fig:other_tasks}
    \vspace{-2mm}
\end{figure*}

\begin{table*}[t]
  \centering
  \setlength{\tabcolsep}{1pt}
  \renewcommand\arraystretch{0.5}
  \small
    \begin{tabular}{c|cccccccccccccccc}
    \toprule[2pt]
          & \small{\textbf{All}}   & \small{Bottle} & \small{Box}   & \small{Bucket} & \small{Cabinet} & \small{Camera} & \small{Cart}  & \small{Chair} & \small{Clock} & \small{Coffee}  & \small{\small{DishWsh.}} & \small{Dispenser} & \small{Door}  & \small{Eyegls} & \small{Fan}   & \small{Faucet} \\ \hline
    \#M   & \textbf{2,346} & 57    & 28    & 36    & 345   & 37    & 61    & 80    & 31    & 55    & 48    & 57    & 36    & 65    & 81    & 84 \\
    \#P   & \textbf{14,068} & 114   & 94    & 74    & 1,174 & 341   & 232   & 1,235 & 106   & 374   & 112   & 162   & 103   & 195   & 172   & 228 \\
    \midrule[1pt]
          &\small{Chair} & \small{Fridge} & \small{Globe} & \small{Kettle} & \small{Keybrd} & \small{Knife} & \small{Lamp}  & \small{Laptop} & \small{Lighter} & \small{MicWav}& \small{Monitor} & \small{Mouse} & \small{Oven}  & \small{Pen}   & \small{Phone} & \small{Pliers} \\
    \hline
    \#M   & 26    & 44    & 61    & 29    & 37    & 44    & 45    & 56    & 28    & 16    & 37    & 14    & 30    & 48    & 17    & 25 \\
    \#P   & 58    & 118   & 130   & 66    & 3,593 & 149   & 165   & 112   & 86    & 85    & 93    & 61    & 214   & 97    & 271   & 59 \\
    \midrule[1pt]
          & \small{Pot}   & \small{Printer} & \small{Remote} & \small{Safe}  & \small{Scissors} & \small{Stapler} & \small{Stcase} & \small{Switch} & \small{Table} & \small{Toaster} & \small{Toilet} & \small{TrashCan} & \small{USB} & \small{Washer} & \small{Window} &  \\
    \hline
    \#M   & 25    & 29    & 49    & 30    & 47    & 23    & 24    & 70    & 101   & 25    & 69    & 70    & 51    & 17    & 58    &  \\
    \#P   & 53    & 376   & 1,490 & 202   & 94    & 69    & 101   & 195   & 420   & 116   & 229   & 208   & 103   & 144   & 195   &  \\
    \bottomrule[2pt]
    \end{tabular}%
  \caption{\textbf{Statistics of PartNet-Mobility Dataset}. \#M and \#P shows the number of models and movable parts respectively.}
  \label{tab:dataset_stats}
  \vspace{-4mm}
\end{table*}%

\subsection{SAPIEN Asset}
SAPIEN Asset is our simulation content, shown in the right box in Figure \ref{fig:sapien_diagram}. 
It contains the large-scale ready-to-simulate PartNet-Mobility dataset, the simulated robot models and scene layouts.

\vspace{-3mm}
\paragraph{PartNet-Mobility Dataset.}
We propose a large-scale 3D interactive model dataset that contains over 14K articulated parts over \textbf{2,346} object models from \textbf{46} common indoor object categories.
All models are collected from 3D Warehouse\footnote{\href{https://3dwarehouse.sketchup.com/}{https://3dwarehouse.sketchup.com/}} and organized as in ShapeNet~\cite{chang2015shapenet} and PartNet~\cite{Mo_2019_CVPR}. 
We annotate 3 types of motions: hinge, slider, and screw, where hinge indicates rotation around an axis (\eg doors); slider indicates translation along an axis (\eg drawers), and screw indicates a combined hinge and slider (\eg bottle caps, swivel chairs). 
For hinge and slider joints, we annotate the motion limit (\ie angles, lengths). 
For screw, we annotate the motion limits and whether the 2 degrees of freedom are coupled. 
Each joint has a parent and a child, and the collection of connected bodies and joints is called an articulation. 
We require the joints of an articulation to follow a tree structure with a single root, since most physical simulator handles tree-structured joint system well. 
Next, for each movable part, we assign a category-specific semantic label. 
Table~\ref{tab:dataset_stats} summarizes the dataset statistics.  Please see the supplementary for more details about the data annotation pipeline.

\newcommand{\twoline}[2]{\begin{tabular}[c]{@{}c@{}}#1\\ #2\end{tabular}}

\vspace{-3mm}
\paragraph{SAPIEN Asset Loader}
Unified Robot Description Format (URDF) is a common format for representing a physical model.  For each object in the SAPIEN Asset, including PartNet-Mobility models and robot models, we provide an associated URDF file, which can be loaded in simulation. For accurate simulation of contact, we decompose meshes into convex parts ~\cite{mamou2016volumetric, huang2018robust}.
We randomize or manually set the physical properties, e.g. friction, damping, density, to appropriate ranges. For robot models, we also provide C++/Python APIs to create a robot piece by piece to avoid complications introduced by URDF.

\subsection{SAPIEN Renderer}
SAPIEN Renderer, shown in the left box of Figure \ref{fig:sapien_diagram}, renders simulated scenes with OpenGL 4.5 and GLSL shaders, which are exposed to the client application for maximal customizability. By default, the rendering module uses a deferred lighting pipeline to provide RGB, albedo, normal, depth, and segmentation from camera space, where lighting is computed with Oren–Nayar diffuse model~\cite{wolff1998improved} and GGX specular model~\cite{walter2007microfacet}. Our customizable rendering interface can suit special rendering needs, and even allow completely different rendering pipelines. We demonstrate this by replacing the fast OpenGL framework with our ray tracer coded with Nvidia OptiX~\cite{parker2010optix} to produce physically accurate images at the cost of rendering time (see Figure~\ref{fig:teaser}).

\subsection{Profiling Analysis}
Our SAPIEN engine can run at about 5000Hz on the manipulation task we will describe in Sec.~\ref{subsec:robotic_manipulation} and can render at about 700Hz with OpenGL mode. 
Tests were performed on a laptop with Ubuntu 18.04, on 2.2 GHz Intel i7-8750 CPU and an Nvidia GeForce RTX 2070 GPU.



%% file: tex/exp.tex
We demonstrate the versatile abilities of our simulator by demonstrating robotic perception and interaction tasks.

\subsection{Robotic Perception}
\label{subsec:perception}
SAPIEN simulator, equipped with the PartNet-Mobility dataset, provides a platform for several robotic perception tasks. 
In this paper, we study the tasks of movable part detection and part motion estimation, which are two important vision tasks supporting downstream robotic interaction.

\begin{table*}[t]
\setlength{\tabcolsep}{1pt}
  \centering
  \renewcommand\arraystretch{1}
    \begin{tabular}{c|c|cccc|ccccc|ccc|cc|c}
    \toprule
    \multicolumn{2}{c|}{} & \multicolumn{4}{c|}{Cabinet}  & \multicolumn{5}{c|}{Table}            & \multicolumn{3}{c|}{Faucet} & \multicolumn{2}{c|}{Fan} & \textbf{All} \\
\hline   Algorithm & Inputs &     \twoline{\small{rot.}}{\small{door}} & \small{body} & \small{drawer} & \twoline{\small{trans.}}{\small{door}} & \small{drawer} & \small{body} & \small{wheel} & \small{door}  & \small{caster} & \small{switch} & \small{base} & \small{spout} & \small{rotor} & \small{frame}  & \textbf{mAP} \\
    \midrule
    \multirow{2}[1]{*}{\twoline{Mask-}{RCNN~\cite{he2017mask}}} & 2D (RGB) & 62.0  & 94.2  & 66.4  & 27.7  & 54.3  & 88.0  & 3.4  & 6.3  & 0.0  & 52.5  & 47.9  & 99.7  & 54.4  & 67.5  & \textbf{53.0}  \\
          & 2D (RGB-D) & 61.7  & 93.0  & 63.0  & 26.3  & 58.6  & 89.9  & 1.4  & 13.2  & 0.0  & 52.1  & 55.8  & 98.9  & 39.4  & 67.4  & \textbf{52.8}  \\ \hline
    \multirow{2}[0]{*}{\twoline{PartNet}{InsSeg~\cite{Mo_2019_CVPR}}} & PC (XYZ) & 20.6  & 65.9  & 35.1  & 9.8  & 15.7  & 71.3  & 1.7  & 1.0  & 0.0  & 34.4  & 55.9  & 64.2  & 50.9  & 74.8  & \textbf{36.1}  \\
          & PC (XYZRGB) & 17.4  & 64.3  & 23.6  & 5.0  & 16.4  & 81.8  & 1.3  & 2.0  & 1.0  & 29.9  & 64.1  & 78.0  & 42.0  & 63.5  & \textbf{37.1}  \\ 
    \bottomrule
    \end{tabular}%
    \caption{\textbf{Movable Part Detection Results.} (AP\% with IoU threshold 0.5) 2D and PC denote 2D images and point clouds as different input modalities for the two algorithms. We show the detailed results for four objects categories and summarize the mAP over all categories. See supplementary for the full table.}
  \label{tab:movable_part_seg}
  \vspace{-4mm}
\end{table*}%

\vspace{-3mm}
\paragraph{Movable Part Detection}
\label{subsubsec:movable_part_detection}
Before interacting with objects by parts, robotic agents need to first detect the parts of interest.
Therefore, we define the task of movable part detection as follows. 
Given a single 2D image snapshot or 3D RGB-D partial scan of an object as input, an algorithm should produce several disjoint part masks associated with their semantic labels, each of which corresponds to an individual movable part of the object.

\begin{figure}[t]
    \centering
    \includegraphics[width=0.8\linewidth]{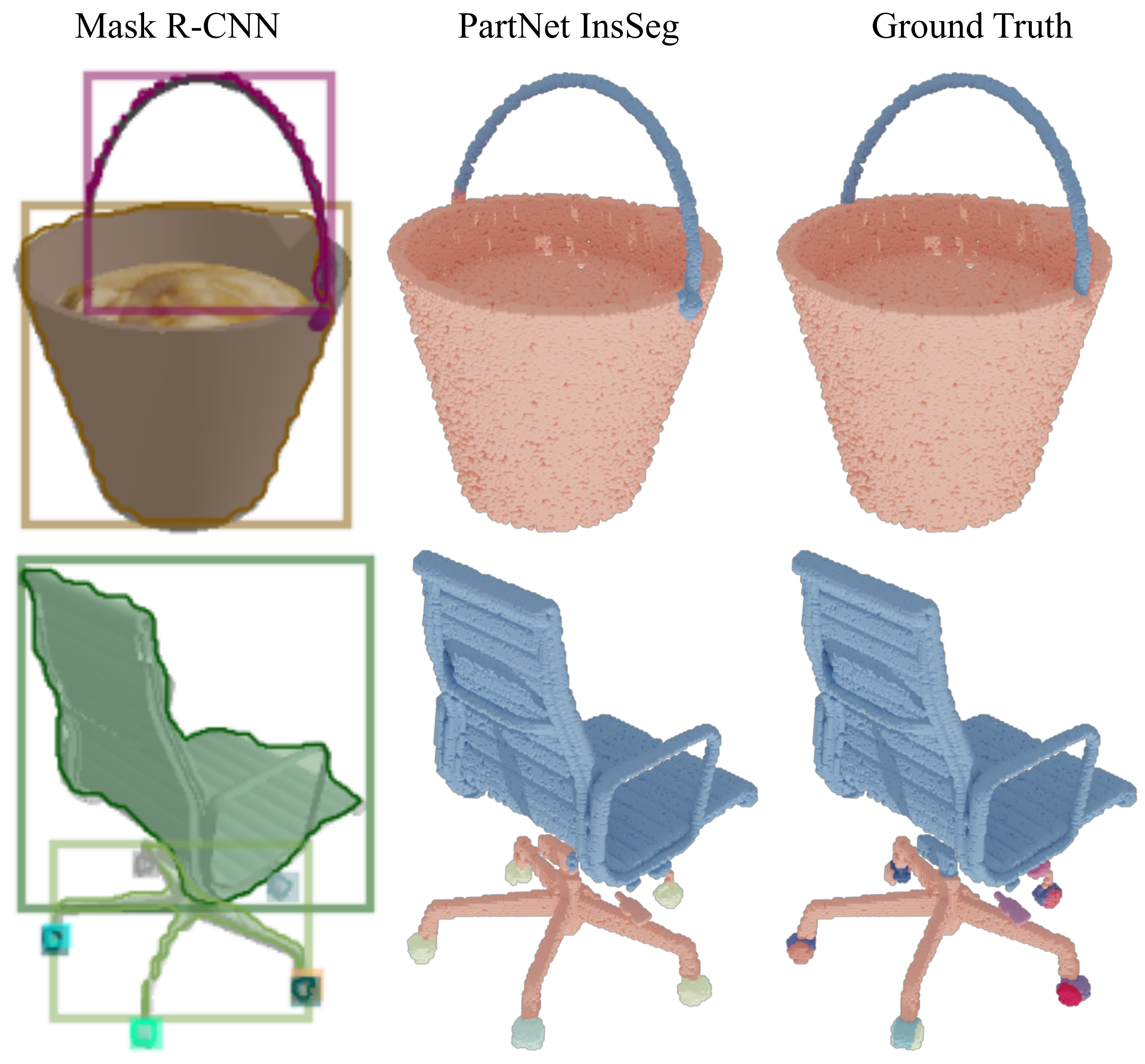}
    \caption{\textbf{Movable Part Detection Results.}  The left column shows the results of Mask R-CNN~\cite{he2017mask}, where each bounding box indicates a detected movable part. The middle and the right columns show the results of PartNet InsSeg~\cite{Mo_2019_CVPR} and the ground truth point clouds respectively, where  different parts are in different color.}
    \label{fig:segmentation}
    \vspace{-4mm}
\end{figure}

Leveraging the rich assets from the  PartNet-Mobility dataset and the SAPIEN rendering pipeline, we evaluate two state-of-the-art perception algorithms for object or part detection in literature. 
Mask R-CNN~\cite{he2017mask} takes a 2D image as input and uses a region proposal network to detect a set of 2D part mask candidates.
PartNet-InsSeg~\cite{Mo_2019_CVPR} is a 3D shape part instance segmentation approach that uses PointNet++~\cite{qi2017pointnet++} to extract geometric features and proposes panoptic segmentation over shape point clouds.

We render each object in the PartNet-Mobility dataset into RGB and RGB-D images from 20 randomly sampled views, with resolution $512\times 512$.
The camera positions are randomly sampled over the upper hemisphere to ensure space coverage. 
Simple ambient and directional lighting without shadows are provided for RGB rendering.
With known camera intrinsics, we lift the 2.5D RGB-D images into 3D partial scans for PartNet-InsSeg experiments.
We use all 2,346 objects over 46 categories from the PartNet-Mobility dataset for this task.
We use 75\% of data (1,772 shapes) for training and 25\% (574 shapes) for testing.
For quantitative evaluation, we report per-part-category Average Precision (AP) scores as commonly used for object detection tasks and average across all part categories to compute mAP for each algorithm.

Table~\ref{tab:movable_part_seg} shows the quantitative results of Mask R-CNN on RGB and RGB-D settings and PartNet-InsSeg on the XYZ (depth-only) and XYZRGB (RGB-D images) settings. We observe that both methods perform poorly on detecting small parts (\eg, table wheel and table caster), and the phenomenon is less severe for object categories that have relatively balanced sizes (\eg, fan and faucet). Small movable parts (\eg, button, switch, and handle) often play critical roles in robot-object interaction, and will demand more well-designed algorithms in the future. 
Figure~\ref{fig:segmentation} visualizes the Mask-RCNN and PartNet-InsSeg part detection results on two example RGB-D partial scans.


\begin{table*}[ht]
\centering
\begin{tabular}{l|l|c|c|c|c|c|c|c}
\toprule
Setting & Algorithm & $H$ acc. & $S$ acc. & $H_o$ err ($m$) & $H_a$ err ($^\circ$)   & $S_a$ err. ($^\circ$) &  door err. ($^\circ$) & drawer err. ($m$)\\
\midrule
RGB-D & ResNet50 & 95.5\% & 95.5\% & 0.168 & 18.9 & 6.35 & 14.4 & 0.0645 \\
RGB-pc & PointNet++ & 95.4\% & 95.5\% & 0.195 & 18.5 & 7.75 & 20.8 & 0.0918 \\
\bottomrule
\end{tabular}
\caption{\textbf{Motion recognition results.} $H$ acc.\ and $S$ acc.\ denotes classification accuracy for hinge and slider respectively. $H_o$ err.\ denotes average distance from predicted hinge origin to ground truth axis. $H_a$/$S_a$ denotes average hinge/slider angle difference from predicted axis to ground truth. door err.\ is average angle difference from predicted door pose to ground truth. drawer err.\ is average length difference from predicted drawer pose to ground truth. }
\label{tab:motion_recognition}
\vspace{-4mm}
\end{table*}

\vspace{-3mm}
\paragraph{Motion Attributes Estimation}
\label{subsubsec:motion_recognition}
Estimating motion attributes for articulated parts gives strong priors for robots before interacting with objects.
In this section, we perform the motion attributes estimation task that jointly predicts the motion type, motion axis, and part state for articulated parts. 

We consider two types of rigid part motions: 3D rotation and translation.
Some parts, such as bottle cap, may have both rotation and translation motions.
For translation motions, we use a 3-dim vector to represent the direction.
For rotation motions, we parameterize the outputs as two 3-dim vectors to specify rotation axis direction and a pivot point on the axis.
We define relative positions of the articulated part with respect to its semantic rest positions as part states.
For example, the rest position for drawers and doors is when they are closed.
However, defining part rest states has intrinsic ambiguities.
For example, round knobs with rotation symmetry do not present a detectable rest position.
Thus, we use a subset of 640 models over 10 categories, which consists of 779 doors and 529 drawers for this task, following the same train and test splits used in the previous section.




We evaluate two baseline algorithms, ResNet-50~\cite{resnet} and PointNet++~\cite{qi2017pointnet++}, that deals with the input RGB-D partial scans using either 2D or 3D formats. 
For ResNet-50, we input RGB-D images augmented with target part mask (5-channel in total).
For PointNet++, we substitute the 5-channel image with its camera-space RGB point cloud.
We train both networks to output a 14-dim motion vector $m=\left(T_r, T_t, p_1^r, p_2^r, p_3^r, d_1^r, d_2^r, d_3^r, d_1^t, d_2^t, d_3^t, x_{\text{door}}, x_{\text{drawer}}\right)$, 
where $T_r$ and $T_t$ respectively output the probability of this joint being rotational and translational, 
$(p_1^r, p_2^r, p_3^r)$ and $(d_1^r, d_2^r, d_3^r)$ indicate pivot point and rotation axis for hinge joints, 
$(d_1^t, d_2^t, d_3^t)$ represents the direction of a proposed slider axis, 
and finally, $x_{\text{door}}$ and $x_{\text{drawer}}$ regress the part poses for doors and drawers respectively.
The part pose is a number normalized within $[0, 1]$ indicating the current joint position.
See supplementary for more details about network architectures, loss designs, and training protocols. 

We summarize the experimental results in Table \ref{tab:motion_recognition}. The classification of different motion types achieves quite high accuracy, and the axis prediction for sliders (translational joints) achieves lower error than for hinges (rotational joints).
In our experiments, ResNet50 achieves better performance than PointNet++. This could be explained by the much higher number of network parameters in ResNet. However, intuition suggests that such 3D information should be more easily predicted directly on 3D data. Future research should focus more on how to improve 3D axis prediction with 3D grounding. 






\subsection{Robotic Interaction}
\label{subsec:robotic_manipulation}

\begin{figure}[t]
    \centering
    \includegraphics[width=0.8\linewidth]{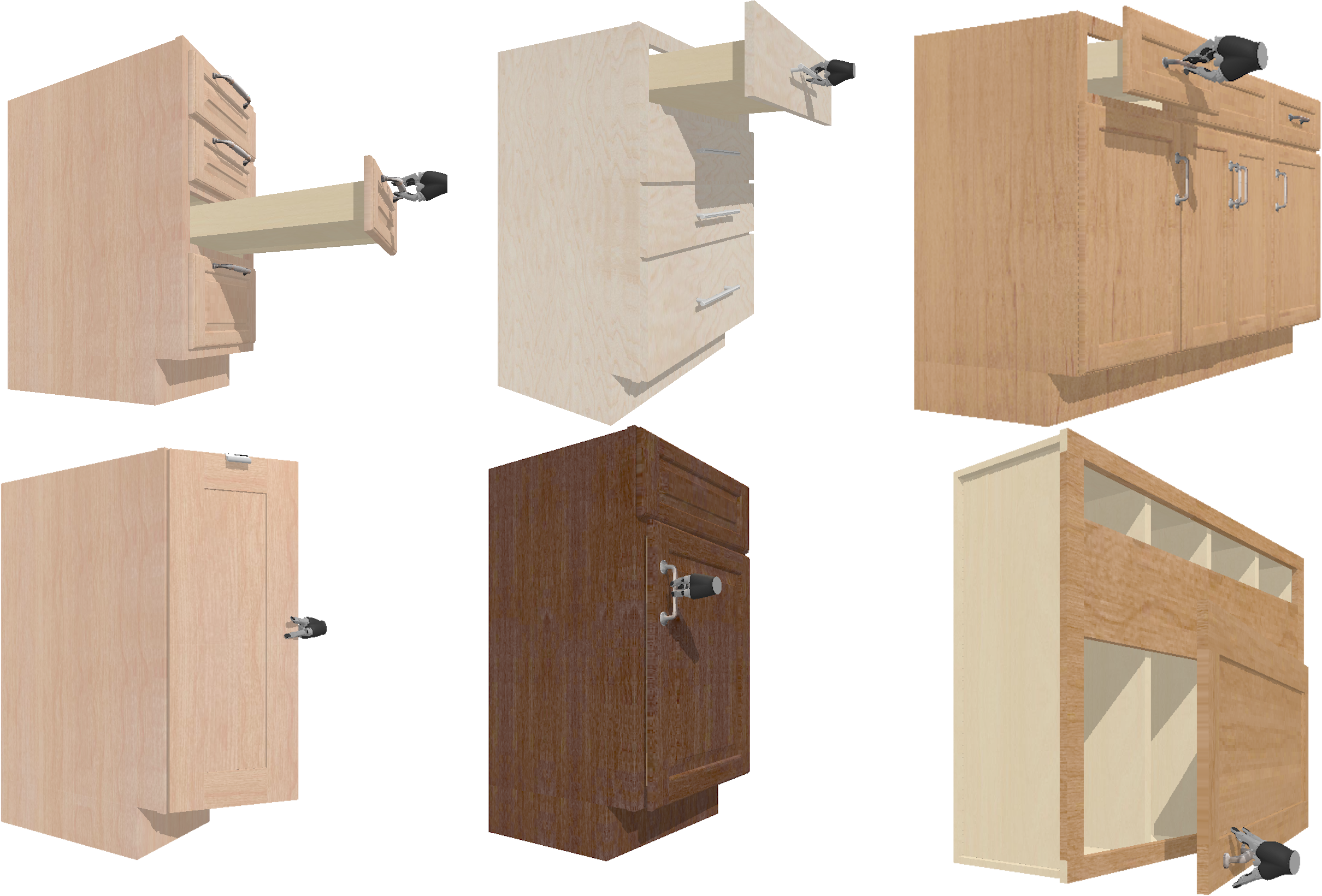}
    \caption{\textbf{Robotic Interaction tasks.} We study two robotic interaction tasks: door-opening and drawer-pulling.}
    \label{fig:manipulation_task}
    \vspace{-4mm}
\end{figure}

With the large-scale PartNet-Mobility dataset, SAPIEN also supports various robotic interaction tasks, including solving low-level control tasks, such as button pushing, handle grasping, and drawer pulling, and planning tasks that require long-horizon logical planning and low-level controls, \eg, removing the mug from a microwave oven and then putting it on a table. Having both diverse object categories and rich intra-class instance variations allows us to perform such tasks on multiple object instances at category levels. Figure~\ref{fig:other_tasks} shows a rich variety of robotic interaction tasks that SAPIEN enables.

In SAPIEN, we enable two modes for robotic interaction tasks: 
1) using perception ground-truth (\eg, part mask, part motion information, and 3D locations) to accomplish the task. In this way, we factor out the perception module and allow algorithms to focus on robotic control and interaction tasks; 
2) using the raw image/point-cloud as inputs, the method needs to develop its own perception, planning and control modules, which is our end-goal for the home-assistant robots to achieve. Also, this mode enables end-to-end learning for perception and interactions (\eg, learning perception with a specific interaction target).

\vspace{-3mm}
\paragraph{Door-opening and Drawer-pulling.}
We perform two manipulation tasks: door-opening and drawer-pulling, as shown in Figure \ref{fig:manipulation_task}. We use a flying gripper (Kinova Gripper 3~\cite{campeau2019kinova}) that can move freely in the workspace. All dynamics properties except gravity, (\eg, contact, friction, and damping) are simulated in our environment. We perform our drawer-pulling tasks on $108$ cabinet instances and door-opening tasks on $77$ cabinet instances.

In our tasks, if the gripper can move a given joint (\eg, slider joint of the drawer, hinge joint of the door) through $90\%$ of its motion range, then it will be regarded as a success. If the agent cannot move the joint to the given threshold or move in the opposite direction, then it fails. The input of the agent consists of point clouds, normal maps and segmentation masks captured by three fixed cameras mounted on the left, right and front of the arena respectively. The agent can also access all information about its self (\eg, 6 DoF pose). 

\vspace{-3mm}
\paragraph{Heuristic Based Manipulation.}
To demonstrate our simulator in manipulation tasks, we first use manually designed heuristic pipelines to solve the tasks. For drawer-pulling, we use point cloud with ground-truth segmentation to detect a valid grasp pose for drawer handle. Then we use velocity controller to pull it to the joint limit. Using ground-truth visual information, we can achieve a $95.3\%$ success rate. 
As for the door-opening task, we first open the door with a small angle using a similar approach (grasp a handle at first). Then we use Position Based Visual Servoing (PBVS) \cite{hutchinson1996tutorial} to track and clamp the edge of the door. Finally, the door is opened by rotating the edge. This method (PBVS) achieves an $81.8\%$ success rate for door opening. A more detailed illustration of this heuristic-based pipeline can be found in our supplementary video.

\vspace{-3mm}
\paragraph{Learning Based Manipulation.}
We also demonstrate the above two tasks using reinforcement learning. We test the generalizability of the RL agent by training on limited objects and testing on unseen objects with different size, density, and motion properties. We adopt Soft Actor-Critic(SAC)~\cite{haarnoja2018soft}, which is one of the SOTA reinforcement learning algorithms, trained on $2, 4, 8, 16$ doors or drawers, and test on the rest unseen models.

We provide three different state representations: 1) raw state of the whole scene (\textbf{raw-exp}), consisting of current positions and velocities of all the parts; 2) mobility-based representation (\textbf{mobility-exp}), with 6D pose of motion axis and average normal, and current joint angles and velocities of the target part; 3) visual inputs (\textbf{visual-exp}), where we set a front-view camera capturing RGB-D images for the object every time step, augmented with segmentation mask for the target part. 

We use the same flying gripper and initialize it on the handle. The grasp pose is generated by the heuristic method as described in the above section. During training, agents receive positive rewards when the target part approaches the joint limit with the opening door/drawer, while obtaining negative rewards when the gripper falls off the handle.
We interact with multiple objects simultaneously during training, and use a shared replay buffer to collect experiences to train SAC. After 1M interaction steps, we evaluate the performance on the unseen objects, each for 20 episodes. 

For doors, the evaluation metric is the average achieved degree. For drawers, we report the success rate of opening 80\% of joint limits. Table~\ref{tab:reinforcement learning} shows our experimental results. For door-opening, the RL agent tends to overfit the training objects, as when the number of training objects grows, the performance drops. However, training on more scenarios will improve the generalization capability with increased test performance. For drawer-pulling, although the performance follows the same pattern as the door, it is relatively stable across the number of training objects. This is because drawers are relatively easier to pull out, as the movement for the gripper almost follows the same pattern every time step. 

Among all the representations, \textbf{mobility-exp} gives the best performance. For doors, \textbf{visual-exp} representation also performs close to \textbf{mobility-exp}; however for drawers, \textbf{raw-exp} is better than \textbf{visual-exp}. This is because the camera is fixed during the interaction. For drawer-opening, the visual features remain almost the same every time step from the front view, so it provides little information about state changing.
These observations lead us to some interesting future work. First, we need proper vision methods to encode the geometric information of the scene, which may change during interaction procedures. Second, although these tasks are not hard for heuristic algorithms, RL-based approaches fail to perform well on all the objects. Future works may study how to enhance the transferability and efficiency of RL on the tasks.

\begin{table}[]
\setlength{\tabcolsep}{1.5pt}
\begin{tabular}{c|c|cccc|cccc}
\toprule
\multicolumn{2}{c|}{Tasks} & \multicolumn{4}{c|}{\twoline{Door}{(Final Angle Degree)}} & \multicolumn{4}{c}{\twoline{Drawer} {(Success Rate)}} \\ \hline
\multicolumn{2}{c|}{}      & 2    & 4    & 8    & 16   & 2    & 4    & 8    & 16    \\ \midrule
raw-exp        & train      & 85.4 & 70.5 & 50.5 & 38.4 &  \textbf{0.84} &  \textbf{0.82}& 0.77 & 0.75      \\
              & test       & 14.7 & 18.7 & 21.2 & 27.3 &  0.61 &  0.63 & 0.66 & 0.66  \\ \hline
mobility-exp       & train      & 88.7 & \textbf{78.6} & \textbf{59.2} & \textbf{41.1} &0.83 & 0.81 &   \textbf{0.79} & \textbf{0.78} \\
              & test       & \textbf{22.9} & \textbf{27.3} & 27.5 &  \textbf{32.8} &     \textbf{0.65} &     \textbf{0.65} &     \textbf{0.69} &      \textbf{0.68} \\ \hline
visual-exp        & train    &    \textbf{90.2} &  65.2&     56.7 &   32.1 & 0.80     & 0.72 &   0.69   &   0.63\\
              & test    & 21.7 &24.5 &  \textbf{28.1} &  29.6 &   0.59   & 0.60     &    0.61  &    0.60   \\ \bottomrule
\end{tabular}
\caption{\textbf{SAC results on door and drawer opening.}}
\label{tab:reinforcement learning}
\vspace{-3mm}
\end{table}

%% file: tex/conclusion.tex
We present SAPIEN, a simulation environment for robotic vision and interaction tasks, which provides detailed part-level physical simulation, hierarchical robotics controllers and versatile rendering options. We demonstrate  that  our  SAPIEN  enables a large  variety  of robotic perception and interaction tasks.




%% file: tex/ack.tex
This research was supported by NSF grant IIS-1764078, NSF grant IIS-1763268, a Vannevar Bush Faculty Fellowship, the Canada CIFAR AI Chair program, gifts from Qualcomm, Adobe, and Kuaishou Technology, and grants from the Samsung GRO program and the SAIL Toyota Research Center.




%% file: supp_tex/appBanno.tex
\begin{figure*}
    \centering
    \includegraphics[width=0.95\linewidth]{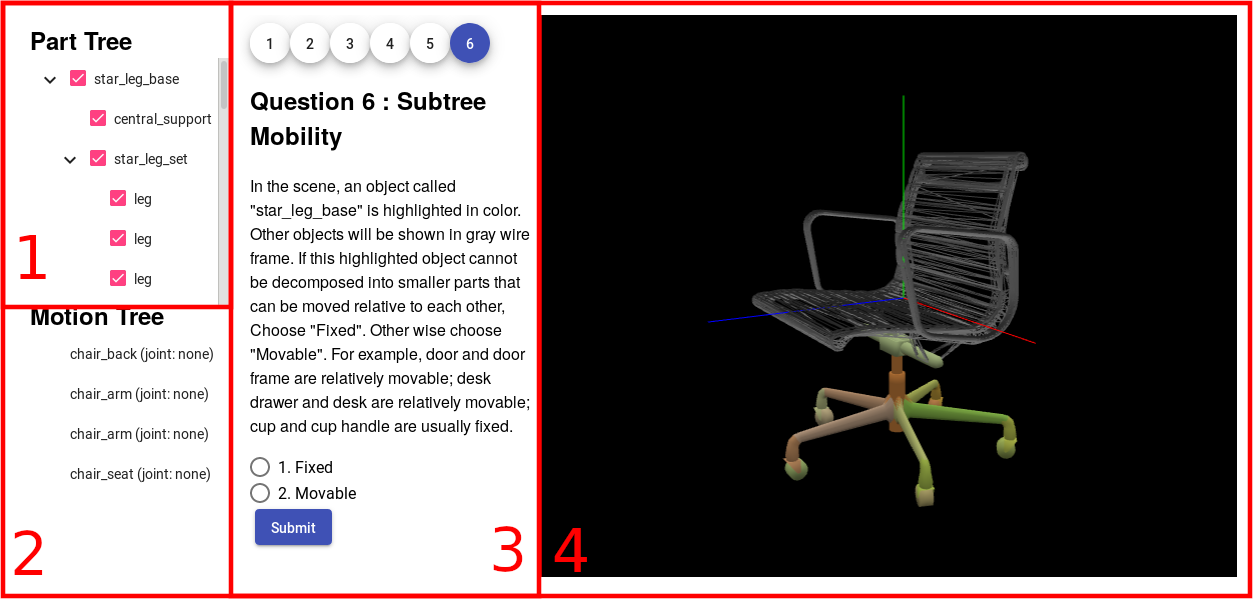}
    \caption{\textbf{Annotation interface.} 1) Part Tree: PartNet semantic tree that proposes fixed parts. 2) Motion tree: annotated movable parts. 3) Question: auto-generated exhaustive questions. 4) Visualization for current question and for motion axis annotation. }
    \label{fig:annotation_interface}
\end{figure*}

We developed a web interface (Figure \ref{fig:annotation_interface}) for mobility annotation. This tool is a question answering (QA) system, which proposes questions based on current stage of annotation. It exploits the hierarchical structures of PartNet to propose objects without relative mobility, and generates new questions based on past annotations. Using this tool, annotators will not miss any movable parts if they answer every question correctly, and they will not face any redundant questions by design. The output mobility annotations are guaranteed to satisfy tree properties suitable for simulation.

The annotation procedure has the following steps:
\begin{itemize}
    \item We start with a PartNet semantic tree, and traverse the tree nodes. Annotators are prompted with questions asking if current subtree has relative motion. If it does not, all parts in this tree will be fixed together; otherwise, the same question is asked again on the child nodes of this subtree.
    
    \item When the PartNet semantic tree traversal is finished, annotators are asked to choose parts that are fixed together.
    
    \item Next, annotators are asked to choose parts that are connected with a hinge (rotational) joint. They will then choose parent-child relation, and annotate axis position/motion limit with our 3D annotation tool.
    
    \item Next, annotators are asked to choose parts that are connected with a slider (translational) joint. They will similarly choose motion parameters and decide if this axis also bears rotation (screw joint).
    
    \item Finally, annotators will annotate each separate object in the scene as ``fixed base'', ``free"", or ``out lier''.
\end{itemize}
The procedure is summarized in the following pseudo-code block.
\begin{algorithm}
    \begin{algorithmic}[1]
        \State Propose fixed parts based on PartNet tree
        \For {There are parts can be fixed together}
        \State Select a group of relatively fixed parts
        \EndFor
        \For {Rotation relationship exists}
        \State Select parent and child
        \State Pick rotation axis
        \State Input motion range
        \EndFor
        \For {Translation relationship exists}
        \State Select parent and child
        \State Pick translation axis
        \State Input motion range / whether it can also rotate
        \EndFor
        \State Choose whether root nodes are fixed/free
    \end{algorithmic}
    \caption*{\textbf{Annotating PartNet-Mobility dataset}}
\end{algorithm}

%% file: supp_tex/appCnet.tex
\newcommand{\twoline}[2]{\begin{tabular}[c]{@{}c@{}}#1\\ #2\end{tabular}}

\begin{table*}[t]
  \scriptsize
  \setlength{\tabcolsep}{0.5pt}
  \centering
    \begin{tabular}{c|c|rrrr|rrrrr|rrr|rrrr|rr|rr}
    \toprule
    & & \multicolumn{3}{c|}{Bottle} & \multicolumn{2}{c|}{Box} & \multicolumn{2}{c|}{Bucket} & \multicolumn{4}{c|}{Cabinet}  & \multicolumn{4}{c|}{Camera}   & \multicolumn{2}{c|}{Cart} & \multicolumn{3}{c}{Chair} \\
\hline   Algorithm & Setting & \multicolumn{1}{c}{tr. lid} & \multicolumn{1}{c}{body} & \multicolumn{1}{c|}{rot. lid} & \multicolumn{1}{c}{rot. lid} & \multicolumn{1}{c|}{body} & \multicolumn{1}{c}{handle} & \multicolumn{1}{c|}{body} & \multicolumn{1}{c}{door} & \multicolumn{1}{c}{body} & \multicolumn{1}{c}{door} & \multicolumn{1}{c|}{drawer} & \multicolumn{1}{c}{lens} & \multicolumn{1}{c}{button} & \multicolumn{1}{c}{body} & \multicolumn{1}{c|}{knob} & \multicolumn{1}{c}{wheel} & \multicolumn{1}{c|}{body} & \multicolumn{1}{c}{wheel} & \multicolumn{1}{c}{seat} & \multicolumn{1}{c}{leg} \\
    \hline
    Mask- & RGB   & 0.0\% & 57.4\% & \multicolumn{1}{r|}{69.3\%} & \multicolumn{1}{r}{49.3\%} & \multicolumn{1}{r|}{65.7\%} & 2.7\% & \multicolumn{1}{r|}{91.7\%} & 62.0\% & \multicolumn{1}{r}{94.2\%} & 27.7\% & \multicolumn{1}{r|}{66.4\%} & \multicolumn{1}{r}{26.7\%} & 20.9\% & 79.0\% & \multicolumn{1}{r|}{4.8\%} & \multicolumn{1}{r}{54.6\%} & \multicolumn{1}{r|}{95.6\%} & \multicolumn{1}{r}{25.1\%} & 97.0\% & 88.3\% \\
          RCNN & RGB-D  & 13.9\% & 68.3\% & \multicolumn{1}{r|}{67.8\%} & \multicolumn{1}{r}{51.5\%} & \multicolumn{1}{r|}{66.5\%} & 1.6\% & \multicolumn{1}{r|}{100.0\%} & 61.7\% & \multicolumn{1}{r}{93.0\%} & 26.3\% & \multicolumn{1}{r|}{63.0\%} & \multicolumn{1}{r}{26.4\%} & 17.0\% & 92.6\% & \multicolumn{1}{r|}{8.1\%} & \multicolumn{1}{r}{55.3\%} & \multicolumn{1}{r|}{93.9\%} & \multicolumn{1}{r}{23.1\%} & 99.0\% & 85.2\% \\
    \hline
    PartNet & XYZ   & 24.5\% & 47.7\% & \multicolumn{1}{r|}{53.5\%} & \multicolumn{1}{r}{27.6\%} & \multicolumn{1}{r|}{46.2\%} & 63.4\% & \multicolumn{1}{r|}{99.7\%} & 20.6\% & \multicolumn{1}{r}{65.9\%} & 9.8\% & \multicolumn{1}{r|}{35.1\%} & \multicolumn{1}{r}{17.0\%} & 0.0\% & 51.4\% & \multicolumn{1}{r|}{0.0\%} & \multicolumn{1}{r}{6.2\%} & \multicolumn{1}{r|}{71.7\%} & \multicolumn{1}{r}{1.2\%} & 93.0\% & 86.4\% \\
          InsSeg & XYZRGB & 5.9\% & 41.3\% & \multicolumn{1}{r|}{54.8\%} & \multicolumn{1}{r}{24.2\%} & \multicolumn{1}{r|}{36.8\%} & 60.7\% & \multicolumn{1}{r|}{98.9\%} & 17.4\% & \multicolumn{1}{r}{64.3\%} & 5.0\% & \multicolumn{1}{r|}{23.6\%} & \multicolumn{1}{r}{10.5\%} & 0.0\% & 46.1\% & \multicolumn{1}{r|}{1.0\%} & \multicolumn{1}{r}{9.4\%} & \multicolumn{1}{r|}{77.3\%} & \multicolumn{1}{r}{1.9\%} & 95.7\% & 89.2\% \\
    \midrule
    & & \multicolumn{3}{c|}{Chair} & \multicolumn{2}{c|}{Clock} & \multicolumn{6}{c|}{CoffeeMachine}            & \multicolumn{2}{c|}{Dishwasher} & \multicolumn{2}{c|}{Dispenser} & \multicolumn{3}{c|}{Display} & \multicolumn{2}{c}{Door} \\
\hline  Algorithm & Setting & \multicolumn{1}{c}{knob} & \multicolumn{1}{c}{caster} & \multicolumn{1}{c|}{lever} & \multicolumn{1}{c}{hand} & \multicolumn{1}{c|}{body} & \multicolumn{1}{c}{button} & \multicolumn{1}{c}{lid} & \multicolumn{1}{c}{body} & \multicolumn{1}{c}{lever} & \multicolumn{1}{c}{knob} & \multicolumn{1}{c|}{container} & \multicolumn{1}{c}{rot. Door} & \multicolumn{1}{c|}{body} & \multicolumn{1}{c}{lid} & \multicolumn{1}{c|}{body} & \multicolumn{1}{c}{rot. screen} & \multicolumn{1}{c}{base} & \multicolumn{1}{c|}{button} & \multicolumn{1}{c}{frame} & \multicolumn{1}{c}{rot. door} \\
    \hline
    Mask- & RGB   & 0.0\% & 2.5\% & \multicolumn{1}{r|}{20.0\%} & \multicolumn{1}{r}{11.4\%} & \multicolumn{1}{r|}{61.4\%} & 14.7\% & 73.4\% & 65.7\% & \multicolumn{1}{r}{0.0\%} & 43.0\% & \multicolumn{1}{r|}{100.0\%} & \multicolumn{1}{r}{70.4\%} & \multicolumn{1}{r|}{90.0\%} & 74.9\% & \multicolumn{1}{r|}{90.1\%} & \multicolumn{1}{r}{74.4\%} & 34.7\% & 0.0\% & 39.4\% & 40.7\% \\
          RCNN & RGB-D  & 0.0\% & 3.6\% & \multicolumn{1}{r|}{13.4\%} & \multicolumn{1}{r}{12.5\%} & \multicolumn{1}{r|}{68.3\%} & 10.4\% & 61.4\% & 67.4\% & \multicolumn{1}{r}{1.0\%} & 35.6\% & \multicolumn{1}{r|}{98.0\%} & \multicolumn{1}{r}{66.8\%} & \multicolumn{1}{r|}{87.8\%} & 73.2\% & \multicolumn{1}{r|}{88.1\%} & \multicolumn{1}{r}{71.3\%} & 33.4\% & 0.0\% & 35.7\% & 54.6\% \\
    \hline
    PartNet & XYZ   & 0.0\% & 1.0\% & \multicolumn{1}{r|}{0.0\%} & \multicolumn{1}{r}{0.0\%} & \multicolumn{1}{r|}{77.0\%} & 0.0\% & 43.6\% & 62.4\% & \multicolumn{1}{r}{0.0\%} & 0.0\% & \multicolumn{1}{r|}{94.0\%} & \multicolumn{1}{r}{50.5\%} & \multicolumn{1}{r|}{67.0\%} & 49.1\% & \multicolumn{1}{r|}{57.6\%} & \multicolumn{1}{r}{66.1\%} & 37.1\% & 0.0\% & 49.2\% & 35.3\% \\
          InsSeg & XYZRGB & 0.0\% & 1.0\% & \multicolumn{1}{r|}{0.0\%} & \multicolumn{1}{r}{0.0\%} & \multicolumn{1}{r|}{79.4\%} & 0.0\% & 81.2\% & 45.8\% & \multicolumn{1}{r}{0.0\%} & 0.0\% & \multicolumn{1}{r|}{85.1\%} & \multicolumn{1}{r}{58.2\%} & \multicolumn{1}{r|}{73.3\%} & 27.4\% & \multicolumn{1}{r|}{39.5\%} & \multicolumn{1}{r}{58.2\%} & 39.1\% & 0.0\% & 34.6\% & 24.6\% \\
    \midrule
    & & \multicolumn{2}{c|}{Eyeglasses} & \multicolumn{2}{c|}{Fan} & \multicolumn{3}{c|}{Faucet} & \multicolumn{2}{c|}{FoldingChair} & \multicolumn{2}{c|}{Globe} & \multicolumn{2}{c|}{Kettle} & \multicolumn{2}{c|}{Keyboard} & \multicolumn{2}{c|}{KitchenPot} & \multicolumn{3}{c}{Knife} \\
\hline    Algorithm & Setting  & \multicolumn{1}{c}{leg} & \multicolumn{1}{c|}{body} & \multicolumn{1}{c}{rotor} & \multicolumn{1}{c|}{frame} & \multicolumn{1}{c}{switch} & \multicolumn{1}{c}{base} & \multicolumn{1}{c|}{spout} & \multicolumn{1}{c}{seat} & \multicolumn{1}{c|}{leg} & \multicolumn{1}{c}{sphere} & \multicolumn{1}{c|}{frame} & \multicolumn{1}{c}{lid} & \multicolumn{1}{c|}{body} & \multicolumn{1}{c}{base} & \multicolumn{1}{c|}{key} & \multicolumn{1}{c}{lid} & \multicolumn{1}{c|}{body} & \multicolumn{1}{c}{blade} & \multicolumn{1}{c}{body} & \multicolumn{1}{c}{blade} \\
    \hline
    Mask- & RGB   & 51.2\% & \multicolumn{1}{r|}{85.2\%} & 54.4\% & 67.5\% & 52.5\% & 47.9\% & \multicolumn{1}{r|}{99.7\%} & 90.6\% & 46.1\% & 98.0\% & \multicolumn{1}{r|}{71.1\%} & \multicolumn{1}{r}{75.2\%} & \multicolumn{1}{r|}{99.4\%} & 15.0\% & \multicolumn{1}{r|}{17.5\%} & \multicolumn{1}{r}{99.0\%} & \multicolumn{1}{r|}{94.5\%} & \multicolumn{1}{r}{11.7\%} & 88.5\% & 33.4\% \\
          RCNN & RGB-D  & 49.2\% & \multicolumn{1}{r|}{84.9\%} & 39.4\% & 67.4\% & 52.1\% & 55.8\% & \multicolumn{1}{r|}{98.9\%} & 93.8\% & 47.2\% & 96.0\% & \multicolumn{1}{r|}{69.6\%} & \multicolumn{1}{r}{94.1\%} & \multicolumn{1}{r|}{100.0\%} & 8.8\% & \multicolumn{1}{r|}{5.1\%} & \multicolumn{1}{r}{100.0\%} & \multicolumn{1}{r|}{95.0\%} & \multicolumn{1}{r}{10.0\%} & 77.8\% & 34.5\% \\
    \hline
    PartNet & XYZ   & 62.1\% & \multicolumn{1}{r|}{93.8\%} & 50.9\% & 74.8\% & 34.4\% & 55.9\% & \multicolumn{1}{r|}{64.2\%} & 91.2\% & 79.4\% & 83.0\% & \multicolumn{1}{r|}{77.6\%} & \multicolumn{1}{r}{71.1\%} & \multicolumn{1}{r|}{74.1\%} & 6.8\% & \multicolumn{1}{r|}{1.0\%} & \multicolumn{1}{r}{94.6\%} & \multicolumn{1}{r|}{94.4\%} & \multicolumn{1}{r}{3.1\%} & 80.1\% & 9.4\% \\
          InsSeg & XYZRGB & 80.6\% & \multicolumn{1}{r|}{92.4\%} & 42.0\% & 63.5\% & 29.9\% & 64.1\% & \multicolumn{1}{r|}{78.0\%} & 86.3\% & 75.6\% & 79.0\% & \multicolumn{1}{r|}{82.0\%} & \multicolumn{1}{r}{87.2\%} & \multicolumn{1}{r|}{90.7\%} & 4.0\% & \multicolumn{1}{r|}{1.0\%} & \multicolumn{1}{r}{93.5\%} & \multicolumn{1}{r|}{95.0\%} & \multicolumn{1}{r}{5.0\%} & 82.7\% & 10.1\% \\
    \midrule
    & & \multicolumn{3}{c|}{Lamp} & \multicolumn{2}{c|}{Laptop} & \multicolumn{4}{c|}{Lighter}  & \multicolumn{3}{c|}{Microwave} & \multicolumn{3}{c|}{Mouse} & \multicolumn{5}{c}{Oven} \\
\hline   Algorithm & Setting & \multicolumn{1}{c}{base} & \multicolumn{1}{c}{rot. bar} & \multicolumn{1}{c|}{head} & \multicolumn{1}{c}{base} & \multicolumn{1}{c|}{screen} & \multicolumn{1}{c}{wheel} & \multicolumn{1}{c}{button} & \multicolumn{1}{c}{body} & \multicolumn{1}{c|}{rot. lid} & \multicolumn{1}{c}{door} & \multicolumn{1}{c}{body} & \multicolumn{1}{c|}{button} & \multicolumn{1}{c}{button} & \multicolumn{1}{c}{wheel} & \multicolumn{1}{c|}{body} & \multicolumn{1}{c}{door} & \multicolumn{1}{c}{knob} & \multicolumn{1}{c}{body} & \multicolumn{1}{c}{tr. tray} & \multicolumn{1}{c}{button} \\
    \hline
    Mask- & RGB   & 54.6\% & 14.6\% & \multicolumn{1}{r|}{64.5\%} & \multicolumn{1}{r}{51.9\%} & \multicolumn{1}{r|}{93.1\%} & 35.0\% & 80.8\% & 96.8\% & 97.0\% & 53.8\% & 94.0\% & 0.0\% & 0.0\% & 46.5\% & \multicolumn{1}{r|}{98.0\%} & \multicolumn{1}{r}{54.0\%} & 49.9\% & \multicolumn{1}{r}{86.8\%} & 1.0\% & 0.0\% \\
          RCNN & RGB-D  & 48.8\% & 10.8\% & \multicolumn{1}{r|}{69.5\%} & \multicolumn{1}{r}{47.2\%} & \multicolumn{1}{r|}{92.8\%} & 57.2\% & 94.1\% & 89.2\% & 92.1\% & 49.5\% & 97.1\% & 0.0\% & 1.0\% & 45.3\% & \multicolumn{1}{r|}{95.2\%} & \multicolumn{1}{r}{53.4\%} & 42.3\% & \multicolumn{1}{r}{93.3\%} & 1.0\% & 0.0\% \\
    \hline
    PartNet & XYZ   & 51.8\% & 8.8\% & \multicolumn{1}{r|}{38.5\%} & \multicolumn{1}{r}{93.0\%} & \multicolumn{1}{r|}{97.7\%} & 1.0\% & 0.0\% & 77.4\% & 80.9\% & 25.9\% & 45.8\% & 0.0\% & 1.0\% & 0.0\% & \multicolumn{1}{r|}{76.0\%} & \multicolumn{1}{r}{23.1\%} & 0.0\% & \multicolumn{1}{r}{36.6\%} & 1.0\% & 0.0\% \\
          InsSeg & XYZRGB & 50.6\% & 9.3\% & \multicolumn{1}{r|}{39.7\%} & \multicolumn{1}{r}{89.8\%} & \multicolumn{1}{r|}{96.1\%} & 9.5\% & 61.4\% & 82.5\% & 84.9\% & 24.3\% & 48.7\% & 0.0\% & 1.0\% & 1.0\% & \multicolumn{1}{r|}{61.1\%} & \multicolumn{1}{r}{26.9\%} & 0.0\% & \multicolumn{1}{r}{49.1\%} & 0.0\% & 0.0\% \\
    \midrule
   & & \multicolumn{3}{c|}{Pen} & \multicolumn{2}{c|}{Phone} & \multicolumn{1}{c|}{Pliers} & \multicolumn{2}{c|}{Printer} & \multicolumn{2}{c|}{Refrigerator} & \multicolumn{2}{c|}{Remote} & \multicolumn{4}{c|}{Safe}     & \multicolumn{1}{c|}{Scissors} & \multicolumn{3}{c}{Stapler} \\
\hline    Algorithm & Setting    & \multicolumn{1}{c}{cap} & \multicolumn{1}{c}{body} & \multicolumn{1}{c|}{button} & \multicolumn{1}{c}{button} & \multicolumn{1}{c|}{base} & \multicolumn{1}{c|}{leg} & \multicolumn{1}{c}{button} & \multicolumn{1}{c|}{body} & \multicolumn{1}{c}{body} & \multicolumn{1}{c|}{door} & \multicolumn{1}{c}{button} & \multicolumn{1}{c|}{base} & \multicolumn{1}{c}{knob} & \multicolumn{1}{c}{button} & \multicolumn{1}{c}{body} & \multicolumn{1}{c|}{door} & \multicolumn{1}{c|}{leg} & \multicolumn{1}{c}{body} & \multicolumn{1}{c}{lid} & \multicolumn{1}{c}{base} \\
    \hline
    Mask- & RGB   & 94.1\% & 91.0\% & \multicolumn{1}{r|}{52.8\%} & \multicolumn{1}{r}{18.4\%} & \multicolumn{1}{r|}{51.4\%} & \multicolumn{1}{r|}{79.9\%} & 2.8\% & \multicolumn{1}{r|}{87.1\%} & \multicolumn{1}{r}{83.0\%} & \multicolumn{1}{r|}{60.7\%} & 35.6\% & 75.2\% & 34.1\% & 0.0\% & 88.5\% & 68.5\% & \multicolumn{1}{r|}{34.2\%} & \multicolumn{1}{r}{32.1\%} & 60.2\% & 84.6\% \\
          RCNN & RGB-D  & 94.1\% & 96.2\% & \multicolumn{1}{r|}{57.6\%} & \multicolumn{1}{r}{12.8\%} & \multicolumn{1}{r|}{50.2\%} & \multicolumn{1}{r|}{78.7\%} & 1.5\% & \multicolumn{1}{r|}{72.3\%} & \multicolumn{1}{r}{81.2\%} & \multicolumn{1}{r|}{55.0\%} & 25.6\% & 78.2\% & 24.5\% & 0.0\% & 92.1\% & 74.6\% & \multicolumn{1}{r|}{57.4\%} & \multicolumn{1}{r}{33.6\%} & 75.0\% & 90.5\% \\
    \hline
    
    PartNet & XYZ   & 67.9\% & 98.0\% & \multicolumn{1}{r|}{53.0\%} & \multicolumn{1}{r}{1.0\%} & \multicolumn{1}{r|}{38.0\%} & \multicolumn{1}{r|}{37.9\%} & 0.0\% & \multicolumn{1}{r|}{34.8\%} & \multicolumn{1}{r}{30.0\%} & \multicolumn{1}{r|}{16.2\%} & 1.0\% & 63.2\% & 0.0\% & 0.0\% & 40.5\% & 30.5\% & \multicolumn{1}{r|}{20.6\%} & \multicolumn{1}{r}{31.7\%} & 49.2\% & 76.7\% \\
          InsSeg & XYZRGB & 15.0\% & 96.2\% & \multicolumn{1}{r|}{25.4\%} & \multicolumn{1}{r}{0.0\%} & \multicolumn{1}{r|}{27.0\%} & \multicolumn{1}{r|}{46.0\%} & 0.0\% & \multicolumn{1}{r|}{48.5\%} & \multicolumn{1}{r}{40.2\%} & \multicolumn{1}{r|}{27.7\%} & 1.0\% & 75.9\% & 0.0\% & 0.0\% & 60.8\% & 42.3\% & \multicolumn{1}{r|}{36.4\%} & \multicolumn{1}{r}{28.5\%} & 83.3\% & 89.5\% \\
    \midrule
   & & \multicolumn{5}{c|}{Suitcase}         & \multicolumn{4}{c|}{Switch}   & \multicolumn{5}{c|}{Table}            & \multicolumn{4}{c|}{Toaster}  & \multicolumn{2}{c}{Toilet} \\
\hline   Algorithm & Setting   & \multicolumn{1}{c}{rot. handle} & \multicolumn{1}{c}{body} & \multicolumn{1}{c}{tr. handle} & \multicolumn{1}{c}{wheel} & \multicolumn{1}{c|}{caster} & \multicolumn{1}{c}{frame} & \multicolumn{1}{c}{lever} & \multicolumn{1}{c}{button} & \multicolumn{1}{c|}{slider} & \multicolumn{1}{c}{drawer} & \multicolumn{1}{c}{body} & \multicolumn{1}{c}{wheel} & \multicolumn{1}{c}{door} & \multicolumn{1}{c|}{caster} & \multicolumn{1}{c}{knob} & \multicolumn{1}{c}{slider} & \multicolumn{1}{c}{body} & \multicolumn{1}{c|}{button} & \multicolumn{1}{c}{lever} & \multicolumn{1}{c}{lid} \\
    \hline
    Mask- & RGB   & 25.5\% & 81.7\% & 74.3\% & \multicolumn{1}{r}{6.2\%} & \multicolumn{1}{r|}{0.0\%} & 85.9\% & 24.3\% & 73.6\% & 60.8\% & 54.3\% & 88.0\% & \multicolumn{1}{r}{3.4\%} & 6.3\% & \multicolumn{1}{r|}{0.0\%} & 40.1\% & \multicolumn{1}{r}{39.0\%} & 90.1\% & 5.9\% & 51.6\% & 98.3\% \\
          RCNN & RGB-D  & 36.4\% & 97.3\% & 70.0\% & \multicolumn{1}{r}{18.1\%} & \multicolumn{1}{r|}{0.0\%} & 74.0\% & 26.0\% & 65.8\% & 22.8\% & 58.6\% & 89.9\% & \multicolumn{1}{r}{1.4\%} & 13.2\% & \multicolumn{1}{r|}{0.0\%} & 40.6\% & \multicolumn{1}{r}{33.0\%} & 94.1\% & 4.0\% & 36.4\% & 98.0\% \\
    \hline
    PartNet & XYZ   & 3.7\% & 53.7\% & 63.6\% & \multicolumn{1}{r}{1.4\%} & \multicolumn{1}{r|}{0.0\%} & 52.3\% & 2.3\% & 4.9\% & 1.0\% & 15.7\% & 71.3\% & \multicolumn{1}{r}{1.7\%} & 1.0\% & \multicolumn{1}{r|}{0.0\%} & 0.0\% & \multicolumn{1}{r}{9.9\%} & 79.3\% & 0.0\% & 0.0\% & 69.3\% \\
          InsSeg & XYZRGB & 4.3\% & 53.2\% & 64.5\% & \multicolumn{1}{r}{2.0\%} & \multicolumn{1}{r|}{0.0\%} & 53.5\% & 1.0\% & 2.1\% & 1.7\% & 16.4\% & 81.8\% & \multicolumn{1}{r}{1.3\%} & 2.0\% & \multicolumn{1}{r|}{1.0\%} & 2.6\% & \multicolumn{1}{r}{20.3\%} & 72.9\% & 0.0\% & 0.0\% & 89.6\% \\
    \midrule
   & & \multicolumn{4}{c|}{Toilet}   & \multicolumn{5}{c|}{TrashCan}         & \multicolumn{3}{c|}{USB} & \multicolumn{4}{c|}{WashingMachine} & \multicolumn{2}{c|}{Window} & \multicolumn{2}{c}{All} \\
\hline    Algorithm & Setting     & \multicolumn{1}{c}{body} & \multicolumn{1}{c}{lid} & \multicolumn{1}{c}{seat} & \multicolumn{1}{c|}{button} & \multicolumn{1}{c}{pad} & \multicolumn{1}{c}{lid} & \multicolumn{1}{c}{body} & \multicolumn{1}{c}{door} & \multicolumn{1}{c|}{wheel} & \multicolumn{1}{c}{rotation} & \multicolumn{1}{c}{body} & \multicolumn{1}{c|}{lid} & \multicolumn{1}{c}{door} & \multicolumn{1}{c}{knob} & \multicolumn{1}{c}{button} & \multicolumn{1}{c|}{body} & \multicolumn{1}{c}{window} & \multicolumn{1}{c|}{frame} & \multicolumn{2}{c}{mAP} \\
    \hline
    Mask- & RGB   & 95.3\% & 64.3\% & 61.1\% & 8.9\% & 43.4\% & 68.1\% & 85.6\% & 35.9\% & 73.7\% & 59.8\% & 65.7\% & 71.3\% & 52.0\% & 6.8\% & 4.4\% & 53.5\% & 55.9\% & 12.2\% & \multicolumn{2}{r}{53.0\%} \\
          RCNN & RGB-D  & 91.8\% & 64.4\% & 62.5\% & 3.0\% & 37.1\% & 69.7\% & 84.9\% & 29.7\% & 69.3\% & 74.4\% & 62.8\% & 68.6\% & 41.4\% & 4.0\% & 0.0\% & 73.3\% & 48.7\% & 13.4\% & \multicolumn{2}{r}{52.8\%} \\
    \hline
    PartNet & XYZ   & 83.2\% & 17.6\% & 1.4\% & 0.0\% & 12.7\% & 67.1\% & 57.7\% & 12.1\% & 5.5\% & 30.0\% & 27.1\% & 22.2\% & 22.4\% & 0.0\% & 0.0\% & 30.5\% & 22.6\% & 83.5\% & \multicolumn{2}{r}{36.1\%} \\
          InsSeg & XYZRGB & 86.5\% & 25.1\% & 5.2\% & 0.0\% & 21.8\% & 75.4\% & 73.5\% & 3.9\% & 5.0\% & 23.9\% & 42.9\% & 12.2\% & 14.5\% & 0.0\% & 0.0\% & 22.9\% & 24.0\% & 85.2\% & \multicolumn{2}{r}{37.1\%} \\
    \bottomrule
    \end{tabular}%
    \caption{\textbf{Movable part segmentation results for all categories}}
  \label{tab:segmentation}%
\end{table*}%

\subsubsection*{Movable Part Segmentation: complete results}
Table \ref{tab:segmentation} shows the movable part segmentation results for all categories in PartNet-Mobility dataset.

\subsubsection*{Motion Recognition: experiment details}
For this task, we normalize the $[0, 2\pi]$ hinge joint range to $[0, 1]$. For sliders, we normalize by the maximum motion range over the dataset to make the motion range prediction within $[0, 1]$.

\paragraph{Algorithm.} The baseline algorithm we use is a a ResNet\cite{resnet} classification and Regression network. The input is the ground truth RGB-D image and the segmentation mask for the target movable part. The output has 7 terms:
    $\hat{T_r}\in \{0,1\}$, whether this part has a rotational joint.\\
    $\hat{T_t}\in \{0,1\}$, whether this part has a translational joint.\\
    $\hat{\mathbf p}^r\in \mathbb R^3$, pivot of a predicted rotational axis.\\
    $\hat{\mathbf d}^{r}\in [-1,1]^3$, direction of a predicted rotational axis.\\
    $\hat{\mathbf d}^{t}\in [-1,1]^3$, direction of a predicted translational axis.\\
    $\hat{x}_{\text{door}}\in [0, 1]$, predicted joint position for a door.\\
    $\hat{x}_{\text{drawer}}\in [0, 1]$, predicted joint position for a drawer.
    
In the following, letters without hat indicates their corresponding ground-truth labels.

In our experiment, we modify the input layer of a ResNet50 network to accept 5 channels, and output layer to output 13 numbers. In addition, we apply tanh activation to produce $\hat{\mathbf d}^{r}, \hat{\mathbf d}^{t}$, and sigmoid activation to produce $\hat{x}_{\text{door}}, \hat{x}_{\text{drawer}}$. The loss has 7 terms:\\
Axis alignment loss, measured by cosine distance:
$$L_{dr} = \sum_{T_r=1} 1 - |\frac{\mathbf d^r\cdot\hat{\mathbf d}^r}{||\mathbf d^r|| ||\hat{\mathbf d}^r||}|
\quad L_{dt} = \sum_{T_t=1} 1 - |\frac{\mathbf d^t\cdot\hat{\mathbf d}^t}{||\mathbf d^t || ||\hat{\mathbf d}^t||}|$$

Pivot loss, measured by the distance from predicted pivot to ground truth joint axis:
$$L_p = \sum_{T_r=1} || \hat{\mathbf p}^r - \mathbf p^r - ((\hat{\mathbf p}^r - \mathbf p^r)\cdot \mathbf d^r)\mathbf d^r ||_2^2$$

Joint type prediction loss:
$$L_{T_r} = -\sum T_r\log \hat{T}_r + (1-T_r)\log (1-\hat T_r)$$
$$L_{T_t} = -\sum T_t\log \hat{T_t} + (1-T_t)\log (1-\hat T_t)$$

Joint position loss, $L_2$ loss between predicted position and ground truth position.
$$L_{\text{door}} = \sum_{\text{valid hinge}}(x_{\text{door}} - \hat{x}_{\text{door}})^2\quad $$

$$L_{\text{drawer}} =\sum_{\text{valid slider}}(x_{\text{drawer}} - \hat{x}_{\text{drawer}})^2 $$

The final loss is a summation of all the losses above:
$$L = L_{dr} + L_{dt} + L_p + L_{T_r} + L_{T_t} + L_{\text{door}} + L_{\text{drawer}}$$
This objective is optimized on mini-batches using proper masking based on $H$ and $S$ values.

\begin{figure*}[t]
 \centering
 \vspace*{1em}
 \includegraphics[width=\linewidth]{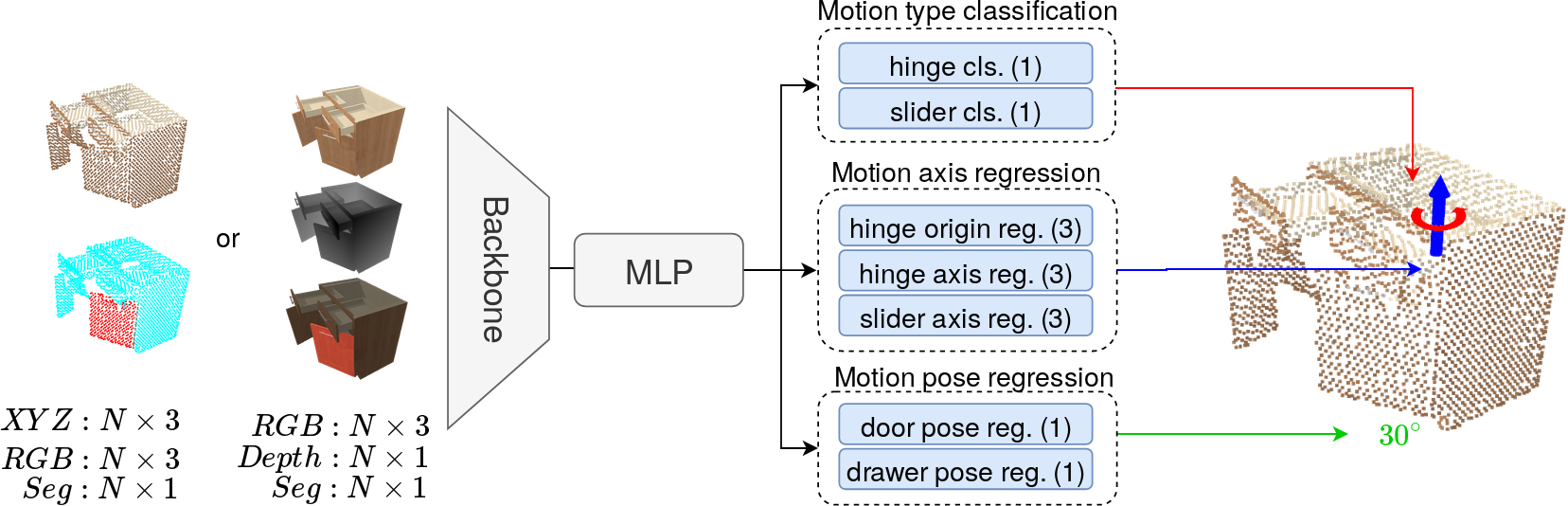}
 \caption{\textbf{Vision Tasks.} Show 2 vision task definitions: inputs + outputs.}
 \label{fig:motion_recognition_net}
\end{figure*}

We repeat this experiment with PointNet++\cite{qi2017pointnet++} operating on 3D RGB-point cloud produced by the same images. For each image, we sample 10,000 points from the partial point cloud (create random copies if the total number of points is less than 10,000). Figure \ref{fig:motion_recognition_net} shows the network structure for the motion recognition tasks.

%% file: supp_tex/appDwrapper.tex
\subsection*{SAPIEN Engine}
\begin{itemize}
    \item \textbf{Articulation:} An articulation is composed of a set of links connected together with transnational or rotational joints \cite{physx}. The most common articulation is a robot.
    \item \textbf{Kinematic/Dynamic joint system: } Both joint systems are an assembly of rigid bodies connected by pairwise constraints. Kinematic system does not respond to external forces while dynamic objects do.
    \item \textbf{Force/Joint/Velocity Controller:} Controller which can control the force/position/velocity of one or multiple joints at once. Like real robot, controller may fail depending on whether the target is reachable.
    \item \textbf{Inertial Measurement Unit(IMU):} A sensor which can measure the orientation, acceleration and angular velocity of the mounted link.
    \item \textbf{Trajectory Controller:} A controller which receive trajectory command and execute to move through the trajectory points. Note that trajectory consist of a sequence of position, velocity and acceleration, while path is simply a set of points without a schedule for reaching each point \cite{hutchinson1996tutorial}.
    \item \textbf{End-effector:} End-effector is a manipulator that performs the task required of the robot, The most common end-effector is gripper.
    \item \textbf{Inverse Kinematics:} Determine the joint position corresponding to a given end-effector position and orientation \cite{siciliano2010robotics}. 
    \item \textbf{Inverse Dynamics:} Determining the joint torques which are needed to generate a given motion. Usualy, the input of inverse dynamics is the output of inverse kinematics or motion planning.
\end{itemize}

\subsection*{SAPIEN Renderer} 
\begin{itemize}
\item \textbf{GLSL} is OpenGL's shading language with describes how the GPU draws visuals.
\item \textbf{Rasterization} is the process of converting shapes to pixels. It is the pipeline used by most real-time graphics applications.
\item \textbf{Ray tracing} is a rendering technique by simulating light-rays, reflections, refractions, etc. It can achieve physically accurate images at the cost of rendering time. OptiX is Nvidia's GPU based ray-tracing framework.
\end{itemize}